\documentclass{article}
\usepackage{geometry}
\PassOptionsToPackage{numbers, sort&compress}{natbib}
\usepackage{natbib}
\usepackage{times}
\usepackage{multirow}
\usepackage[utf8]{inputenc} % allow utf-8 input
\usepackage[T1]{fontenc}    % use 8-bit T1 fonts
\usepackage[colorlinks=true, citecolor=green]{hyperref}       % hyperlinks
\usepackage{url}            % simple URL typesetting
\usepackage{subcaption}
\usepackage{fullpage}
\usepackage{authblk}
\usepackage{longtable}
\usepackage{booktabs}
\usepackage{verbatim}
\usepackage{lscape}

\usepackage{amssymb}
\usepackage{nicefrac}       % compact symbols for 1/2, etc.
\usepackage{amsfonts}       % blackboard math symbols
\usepackage{amsmath}

\usepackage{mathtools}
\usepackage{siunitx}
\usepackage{setspace}

\makeatletter
\def\set@curr@file#1{\def\@curr@file{#1}} %temp workaround for 2019 latex release
\makeatother

\begin{document}

\newcommand{\titlephrase}{LLMs Understand Glass-Box Models, Discover Surprises, \\ and Suggest Repairs}
\newcommand{\packagelink}{\href{https://github.com/interpretml/TalkToEBM}{\texttt{github.com/interpretml/TalkToEBM}}}
\newcommand{\packagelinkshort}{\href{https://github.com/interpretml/TalkToEBM}{\texttt{TalkToEBM}}}

\title{\titlephrase}

\author[1,2]{Benjamin J. Lengerich\thanks{\texttt{blengeri@mit.edu}}}
\author[3]{Sebastian Bordt\thanks{\texttt{Work done while at Microsoft Research.}}}
\author[4]{Harsha Nori}
\author[5]{Mark E. Nunnally}
\author[5]{Yin Aphinyanaphongs}
\author[1,2]{Manolis Kellis\thanks{\texttt{manoli@mit.edu}}}
\author[4]{Rich Caruana\thanks{\texttt{rcaruana@microsoft.com}}}

\affil[1]{Massachusetts Institute of Technology}
\affil[2]{Broad Institute of MIT and Harvard}
\affil[3]{University of Tübingen, Tübingen AI Center}
\affil[4]{Microsoft Research}
\affil[5]{NYU Langone Health}

\maketitle

\begin{abstract}
We show that large language models (LLMs) are remarkably good at working with interpretable models %, especially glass-box models
that decompose complex outcomes into univariate graph-represented components. 
By adopting a hierarchical approach to reasoning, LLMs can provide comprehensive model-level summaries without ever requiring the entire model to fit in context. 
This approach enables LLMs to apply their extensive background knowledge to automate common tasks in data science such as detecting anomalies that contradict prior knowledge, describing potential reasons for the anomalies, and suggesting repairs that would remove the anomalies. 
We use multiple examples in healthcare to demonstrate the utility of these new capabilities of LLMs, with particular emphasis on Generalized Additive Models (GAMs). 
Finally, we present the package \packagelinkshort~as an open-source LLM-GAM interface. 
\end{abstract}

\begin{figure}[t]
    \centering
    \includegraphics[width=0.9\textwidth]{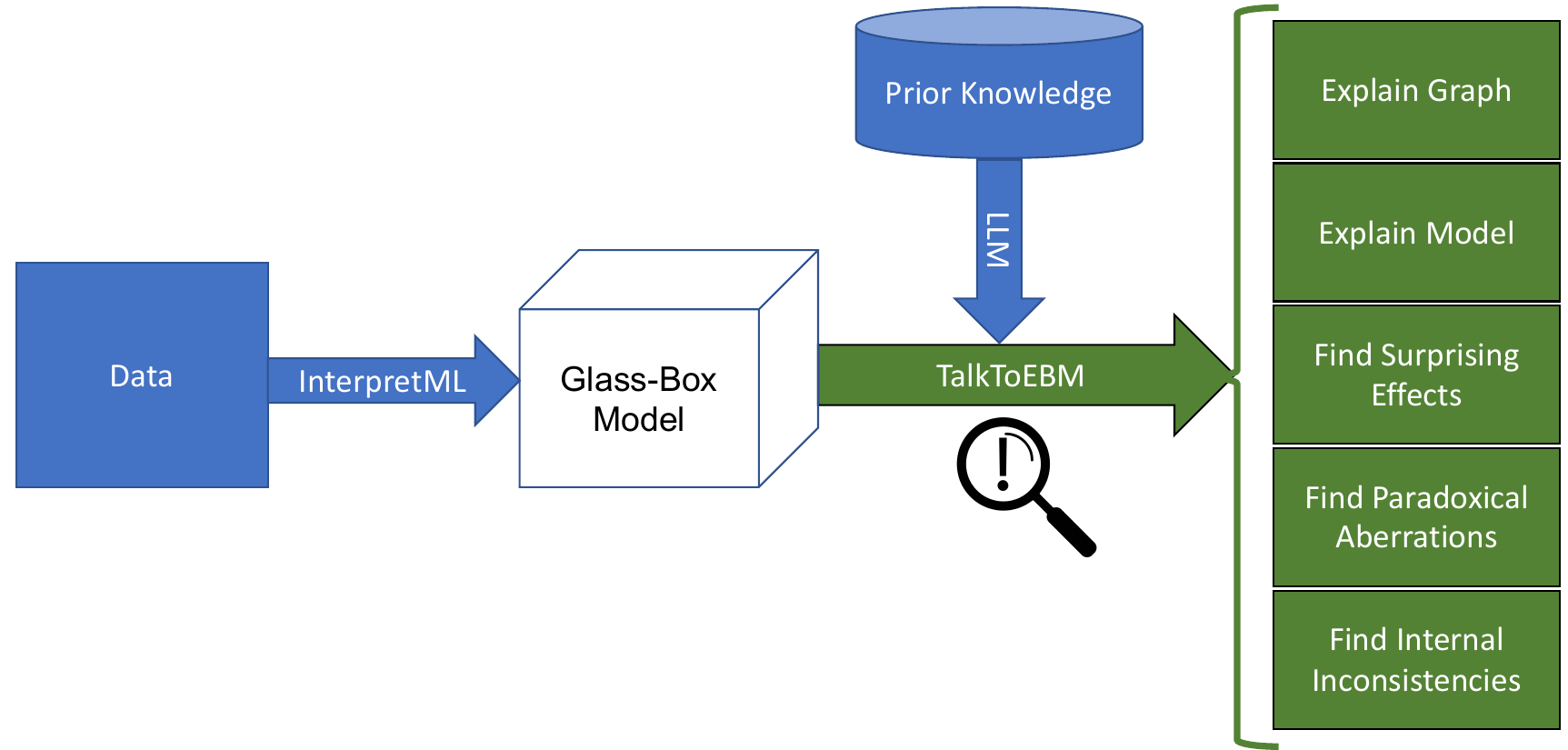}
    \caption{Our approach to connect LLMs with data interpretations. By using glass-box interpretable models that can be decomposed into modular components, we enable the LLM to use hierarchical reasoning by considering each modular component in turn. This enables the LLM to analyze large models and perform complex reasoning tasks without requiring the entire model to be stored in a single context window.}
    \label{fig:graphical_abstract}
\end{figure}

\section*{Introduction}
Large language models (LLMs) offer the potential to automate data science through natural language interfaces, but it is difficult to embed complex models or datasets in confined context windows. 
While GPT-4 has a context window size of up to 32k tokens, paying equal attention to all parts of the context remains a challenge \cite{liu2023lost} and the practicality of lengthy context windows is questionable. 
Machine learning models often involve billions of parameters, accentuating the need for compact, modular function representations that more easily interface with LLMs. 

In this paper, we show that LLMs pair remarkably well with interpretable models that are decomposable into modular components.
Specifically, we show that GPT-4 is able to describe, interpret and debug univariate graphs, and by applying a form of chain-of-thought reasoning\cite{wei2022chain}, GPT-4 can understand Generalized Additive Models (GAMs). 
GAMs \cite{hastie1990generalized,caruana2015intelligible} represent complex outcomes as sums of univariate component functions (graphs); thus, by analyzing each of these component functions in turn, the LLM does not need to understand the entire model at once. 
After analyzing and summarizing each graph, the LLM can operate on component summaries to produce model-level analyses. 
This modularity simplifies the application of LLMs to data science and machine learning and enables LLM-based analyses to scale to very large datasets while staying within small context windows. 

Next, we show that because LLMs have large amounts of prior knowledge, LLMs can be used to provide model interpretations that are grounded in domain expertise. 
Specifically, we show that LLMs can automatically detect surprises and anomalies in models such as GAM graphs that appear to contradict expectations. 
By highlighting these surprises, the LLM can suggest problems in all aspects of the analysis, including data collection, imputation, model fitting, or model specification. 
This is critical for good data science because real-world datasets are universally polluted by confounders and systemic biases that are often obvious only after close examination with domain expertise (e.g. treatment effects in healthcare \cite{lengerich2022death}); LLMs with extensive prior knowledge offer a system to automatically detect and report these potential problems. 
Depending on the application, these potential problems may require users to correct the data, correct the model, or change the underlying system. 

Ultimately, LLMs offer the potential to be used as tools that automate important yet repetitive aspects of the data science process. 
LLMs are especially useful when paired with statistical tools, such as glass-box interpretable models, that break complex reasoning tasks into separable components. 
While LLMs such as GPT-4 are not yet able to directly understand large tables of tabular data, they are able to interface with glass-box GAMs trained on that data, providing a new opportunity for humans to interact and learn with their data.

\subsection*{Our Approach}
Our approach (Figure \ref{fig:graph_to_text}) is to use model representations that provide \emph{separable} components. 
Separable components can be independently analyzed, summarized, and combined with hierarchical reasoning.\footnote{This approach of first using component-level prompts before using model-level prompts is similar to chain-of-thought reasoning \cite{wei2022chain} which attempts to improve LLM reasoning by asking the LLM to explicitly work through a problem, progressively solving more difficult reasoning tasks.}
Glass-box models (e.g. GAMs) that can be decomposed into univariate component graphs fit this approach because the univariate components can be separated and later combined without approximation. 
Thus, we use GAMs to split the complex task of reasoning about an entire model into a sequence of small tasks of reasoning about individual graphs. 
Concretely, we prompt the LLM with a general introductory prompt about the task, model and data set, then successively provide each graph as key-value lists. 
For each graph, we can ask the LLM to perform various tasks (for example, to summarize and answer questions about the graph). 
To draw conclusions about the entire model, the summaries of individual graphs are aggregated in a new query to the LLM. 
This approach is available in our open-source software package \packagelinkshort.

\subsection*{Related Work}
The disruptive potential of using LLMs to automate tasks in data science has inspired several related approaches. 
Slack et al. \cite{slack2022talktomodel} develop a natural language interface to give practitioners conversational access to model explanations. 
Their work intends to provide access to generic (potentially black-box) models, and so the LLM does not have direct access to model internals. 
As a result, \cite{slack2022talktomodel} cannot use the same chain-of-thought approach we explore with GAMs to enable scalability and complex model-level reasoning. 

On the other end of the pipeline for automated data science, Bisercic et al. \cite{bisercic2023interpretable} showed that LLMs can extract tabular datasets from unstructured text and then train interpretable models (e.g. linear regression and small decision trees) on top of this data. 
Similarly, recent works have explored the potential of LLMs for data wrangling and cleaning \cite{narayan2022can,vos2022towards}, or traditional supervised and unsupervised learning tasks like classification or density estimation \citep{borisov2022language,hegselmann2023tabllm,wang2023anypredict,onishi2023tabret}. 
These works usually rely on fine-tuning, suggesting that today's LLMs have only a limited ability to solve complex tasks with raw tabular data in-context. 
All in all, these approaches to automated data preprocessing and model estimation are complementary systems to the analytical system proposed in this paper, together suggesting a path toward fully automated data science which includes automated data preprocessing, model fitting, and interpretations. 

\paragraph{The Importance of Iterative Investigations in Data Science}
While the predictive accuracy of machine learning tools is undoubtedly crucial, it is not sufficient as a standalone objective when employing these tools to analyze and influence real-world systems. 
Instead of singularly pursuing accuracy, we need to embark on a cyclical investigative process to comprehend the model's nuances, the effects it has learned, and the implications it holds for the dataset and corresponding real-world system.
Taking healthcare data as an instance, we may develop a system that effectively predicts the outcomes for hospitalized patients. However, applying this retrospectively derived knowledge to future diagnostic policies can lead to flawed treatment decisions. 
This occurs because the patients identified as low-risk in the retrospective study are those who received effective treatment, not necessarily those who inherently bore the least risk \cite{caruana2015intelligible}.
Moreover, relying heavily on predictive accuracy can be deceptive, particularly considering the inconsistent recording of real-world healthcare data \cite{boag2022ehr}, which could lead to statistical endorsement of unrelated confounding factors \cite{zech2018variable}. Hence, using predictive accuracy as an end goal, our goal is to employ interpretable models to uncover unexpected effects. 
This mirrors the philosophy proposed by \cite{lengerich2022death}, which demonstrated that systemic confounding presents opportunities to enhance healthcare. 
This prior work relied on human experts to design explicit statistical tests for each category of surprising pattern in GAM feature graphs, providing a handle on false-discovery rates but limiting the breadth of potential applications. 
Our work builds on this insight that interpretable models reveal hidden confounding and proposes to automate surprise discovery with LLMs.

\section*{Results}
We begin by examining the ability of LLMs to understand individual graphs. 
We next outline how this capability can be used to explain entire models and summarize data sets. 
Finally, we use the domain knowledge embedded in the LLM to find surprising effects (i.e. hidden confounding). 
We are particularly interested in the use of this approach to identify confounding factors and suboptimal treatment protocols in observational healthcare data; thus, in all of these experiments, we use a dataset of pneumonia patients from \cite{cooper1997evaluation} as a running example. 
The approach is broadly applicable and more examples on publicly available datasets are included in the package repository \packagelinkshort.

\begin{figure}[htp]
    \centering
    \includegraphics[width=\textwidth]{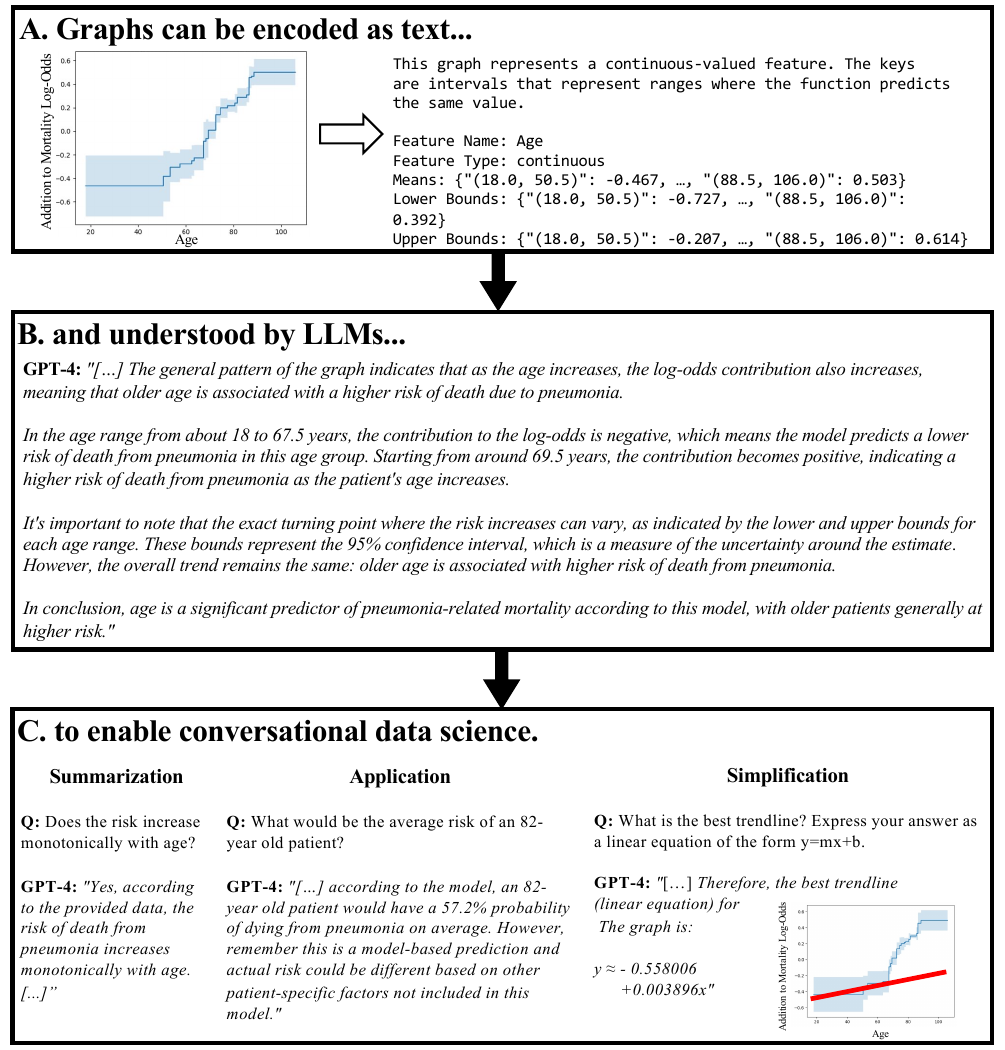}
    \caption{Graphs can be encoded into JSON text and understood by an LLM.
    GPT-4 successfully understands the graph and enables conversational dialogue to speed up data science.
    }
    \label{fig:graph_to_text}
\end{figure}

\subsection*{LLMs Understand Graphs}
LLMs are able to understand and reason about simple graphs. 
The main challenge is to encode each graph in a text description which enables straightforward reasoning. 
Piecewise-constant graphs, such as those estimated by tree-based GAMs, can be exactly described as a list of regions and values. 
By encoding piecewise-constant graphs as JSON objects of x-bins and y-values (Figure~\ref{fig:graph_to_text}A), we pass an exact and efficient description of the graph to the LLM. 
The LLM can summarize this encoding (Figure~\ref{fig:graph_to_text}B) and enter into a dialogue with the user (Figure~\ref{fig:graph_to_text}C). 

For example, consider the following question about the influence of age on pneumonia mortality risk (Figure~\ref{fig:graph_to_text}A):

\begin{quote}{\bf Question:} {\it What would be the average risk of a 82-year old patient?}\end{quote}
\begin{quote}{\bf GPT-4:} {\it "The average risk for an 82-year old patient can be derived from the mean log-odds contribution for the age interval "(81.5, 85.5)", which is 0.29. 

Log-odds are a way of expressing probability and can be converted into a probability with the following formula: p = exp(log-odds) / (1 + exp(log-odds)). 

Using this formula, the average risk is approximately 57.2\%. 

This means that, according to the model, an 82-year old patient would have a 57.2\% probability of dying from pneumonia on average. However, remember this is a model-based prediction and actual risk could be different based on other patient-specific factors not included in this model."}
\end{quote}

This answer is correct since the relevant interval in the graph is indeed {\tt (81.5, 85.5): 0.29} and the arithmetic translation from log-odds to probability is correct. 
The LLM can also do simple reasoning about this graph:

\begin{quote}{\bf Question:} {\it Does the risk increase monotonically with age?}\end{quote}
\begin{quote}{\bf GPT-4:} {\it "Yes, according to the provided data, the risk of death from pneumonia increases monotonically with age. [...]"}
\end{quote}

\begin{quote}{\bf Question:} {\it Which patients are most at risk?}\end{quote}
\begin{quote}{\bf GPT-4:} {\it "According to the model's output, the patients most at risk are those in the oldest age group, specifically those aged 88.5 to 106 years.  [...]"}
\end{quote}

More complicated analysis questions begin to push at the bounds of GPT-4's reasoning capabilities:
\begin{quote}{\bf Question:} {\it What is the average increase in risk due to a 5-year increase in age?}\end{quote}
\begin{quote}{\bf GPT-4:} {\it "To calculate the average increase in risk due to a 5-year increase in age, we need to look at the mean log-odds contributions of adjacent 5-year age groups and calculate the difference. However, the age groups provided are not all of equal length, so we cannot calculate an exact average increase in risk for a 5-year increase in age. 

If we wanted to estimate this, we could consider the log-odds difference between the "(77.5, 80.5)" and "(81.5, 85.5)" age groups. The mean log-odds contribution increases from 0.216 to 0.29, a difference of 0.074 in log-odds.

To convert this difference in log-odds to a difference in probability, we would use the formula: p = exp(log-odds) / (1 + exp(log-odds)). 

Applying this formula, we find that a 5-year increase in age from 80 to 85 years is associated with an increase in the average risk from about 53.8\% to 57.2\%, which is about a 3.4\% increase in risk due to a 5-year increase in age.

Bear in mind that this is an approximation based on the provided data, and the actual increase in risk may vary depending on the specific age range and other factors. Also, as previously mentioned, there are many other factors that can influence a patient's risk of dying from pneumonia, not just age."}
\end{quote}

This answer from GPT-4 is factually correct but incomplete. 
We asked for the average increase in risk represented by the graph, but GPT-4 opted to answer only about two local averages. 
This example demonstrates the common need for LLMs to perform guided bottom-up analyses rather than unguided global queries, motivating our strategy to interface with the LLM via modular function representations rather than non-separable black-box models.  

\subsection*{LLMs Understand Interpretable Models}

\begin{figure}[htp]
    \centering
    \includegraphics[width=0.85\textwidth]{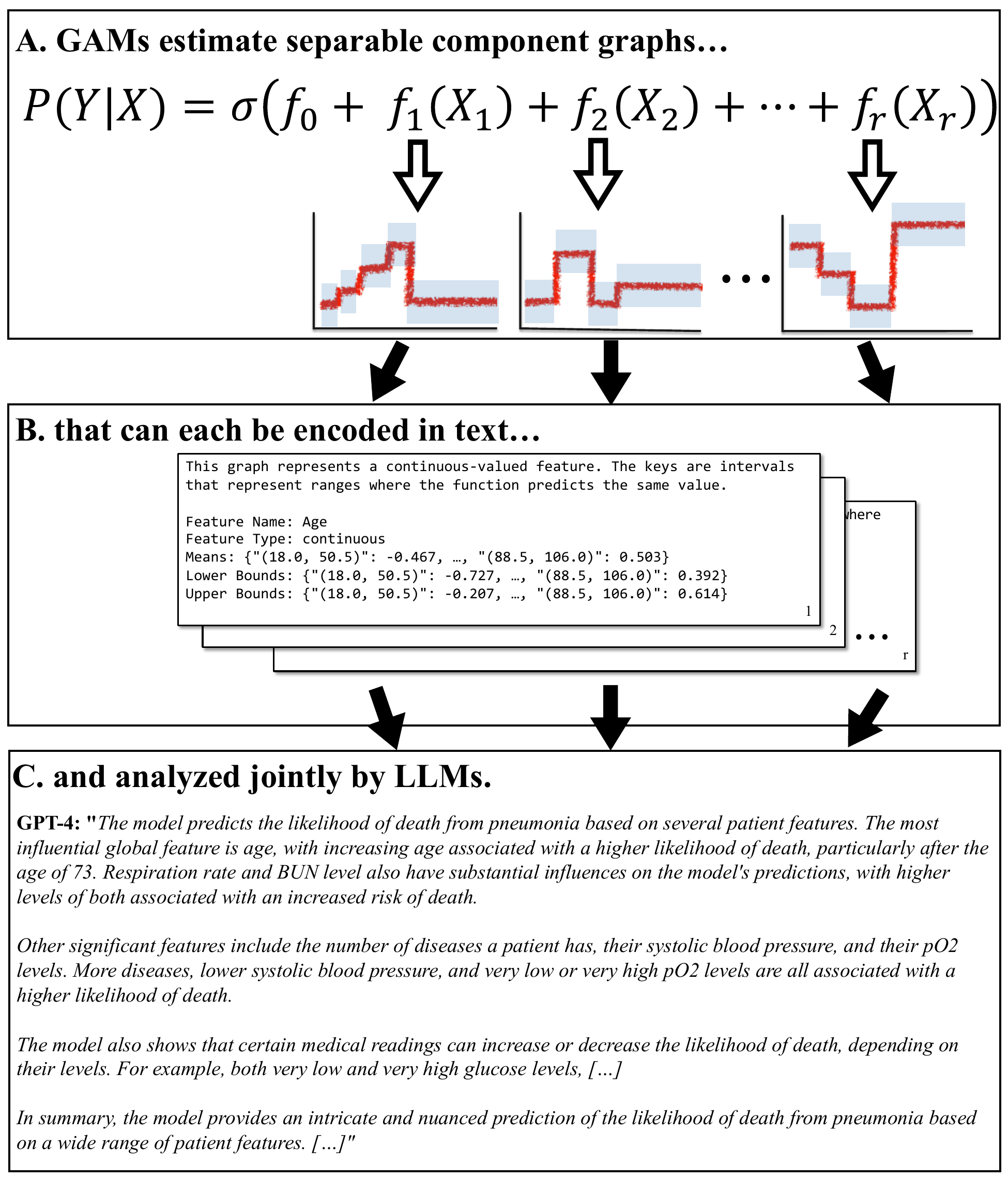}
    \caption{LLMs can be used to reason about complicated outcomes if the outcomes are first broken down into an additive model before passing to the LLM.}
    \label{fig:model}
\end{figure}

GAMs are separable into a set of univariate component functions (Figure~\ref{fig:model}A). 
Starting from the univariate component graphs, we use chain-of-thought prompting \cite{wei2022chain} to summarize the effects of the different features in the model. 
For example, in the pneumonia dataset, the model summarizes the role of `respiration rate' as follows:

\begin{quote}
{\bf Question: } {\it "Now provide a brief executive summary of the role of the feature in the model. Assume that the reader already knows what the model is about, so you don't need to describe the outcome or feature details. Use at most 7 sentences. Avoid boilerplate sentences like 'other factors may also impact this likelihood' or 'further analysis would be needed'."} 

{\bf Respiration Rate (Feature Summary):} "\it The model's output indicates an increasing likelihood of death as the respiration rate increases. Specifically, a respiration rate of 32 or less tends to decrease the likelihood of patient death, while a rate above 32 increases this likelihood. However, the model exhibits some uncertainty in its predictions, as shown by overlapping confidence intervals across different respiration rate intervals. This suggests that while the respiration rate is influential, its precise impact may vary between patients."   
\end{quote}

We then aggregate the different summaries and ask the LLM to provide an overall summary of the GAM. We can also provide the model with additional information. 
Because GAMs also provide straightforward measures of feature importance, we also provide the LLM with global feature importance scores. 

\begin{quote}
{\bf Model Summary (GPT-4, Pneumonia):} {"\it The model predicts the likelihood of death from pneumonia based on several patient features. The most influential global feature is age, with increasing age associated with a higher likelihood of death, particularly after the age of 73. Respiration rate and BUN level also have substantial influences on the model's predictions, with higher levels of both associated with an increased risk of death.

Other significant features include the number of diseases a patient has, their systolic blood pressure, and their pO2 levels. More diseases, lower systolic blood pressure, and very low or very high pO2 levels are all associated with a higher likelihood of death.

The model also shows that certain medical readings can increase or decrease the likelihood of death, depending on their levels. For example, both very low and very high glucose levels, pH levels, and potassium levels can increase the risk of death, while moderate levels are associated with a lower risk.

Finally, the model shows a considerable amount of uncertainty in its predictions for certain features, such as age, heart rate, and albumin levels. This is likely due to a lack of data points in certain ranges or the complex, non-linear relationships these features may have with the outcome.

In summary, the model provides an intricate and nuanced prediction of the likelihood of death from pneumonia based on a wide range of patient features. However, given the complexity of the relationships and the uncertainty in some predictions, the model's output should be interpreted with caution and used in conjunction with clinical judgement."}
\end{quote}

\subsection*{LLMs Automatically Find Surprising Effects}
\begin{figure}[htp]
    \centering
    \includegraphics[width=\textwidth]{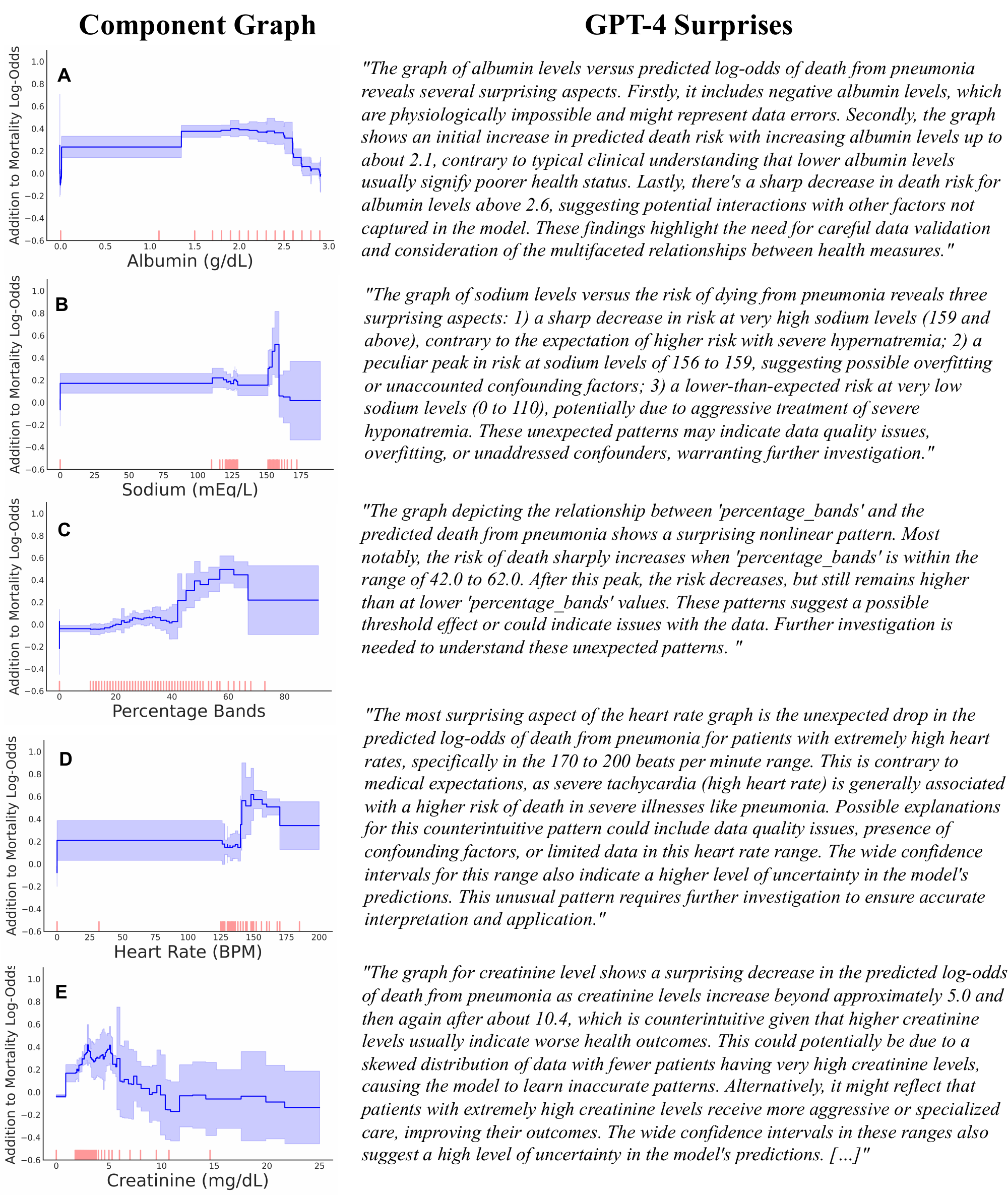}
    \caption{Pneumonia effects and surprises. In each example, we show the estimated impact on patient mortality (left) with 95\% confidence intervals shaded and observed training data points as red ticks along the horizontal axis, and (right) the surprises reported by GPT-4.}
    \label{fig:pneumonia_all}
\end{figure}

Finally, LLMs can apply their embedded domain knowledge to provide grounded interpretations of these graphs and models. 
We are particularly motivated to use LLMs to automatically detect surprising effects, as this is a common but difficult task in data analysis that typically requires domain expertise and iteratative data exploration, model fitting, and evaluation. 
However, LLMs can automate this process by generating a ranked list of surprises that may indicate underlying model or data problems. 
To accomplish this, we include in our system prompt an instruction to pay attention to abnormal and surprising effects:
\begin{quote}
    {\bf System Prompt:} "\it [...] Pay special attention to values or patterns that appear abnormal. It is very important that you alert the user about these potentially surprising aspects of the graphs."
\end{quote}

Once the LLM has been prompted to save notes about these surprises, we can subsequently ask the LLM to summarize and rate the different surprises.
Surprises can be identified in individual graphs:
\begin{quote}{\bf Question:} {\it "Great, now summarize the the most important surprising aspects of this graph in at most 5 sentences."} \end{quote}
\begin{quote}{\bf GPT-4:} {\it "The graph for creatinine level shows a surprising decrease in the predicted log-odds of death from pneumonia as creatinine levels increase beyond approximately 5.0 and then again after about 10.4, which is counterintuitive given that higher creatinine levels usually indicate worse health outcomes. This could potentially be due to a skewed distribution of data with fewer patients having very high creatinine levels, causing the model to learn inaccurate patterns. Alternatively, it might reflect that patients with extremely high creatinine levels receive more aggressive or specialized care, improving their outcomes. The wide confidence intervals in these ranges also suggest a high level of uncertainty in the model's predictions. These surprising patterns warrant further investigation into the data and consultation with clinical experts."}\end{quote}

By asking the model to summarize the entire model as a ranked list of surprises, we generate an ordered list of hypotheses to investigate for confounding or novel discoveries (Figure~\ref{fig:pneumonia_all}). 

\begin{quote}{\bf Finding Surprises and Abnormal Patterns (GPT-4, Pneumonia):} {\it "The Generalized Additive Model (GAM) predicts the likelihood of death from pneumonia based on patient factors. However, it reveals several surprising patterns.

The most surprising factors are Heart Rate, Systolic Blood Pressure, Albumin Level, Percentage Bands, Creatinine Level, and Sodium Level. 

The Heart Rate graph shows a decrease in predicted risk of death for patients with extremely high heart rates, contrary to medical expectations. 

The Systolic Blood Pressure graph shows a counterintuitive pattern: as blood pressure increases, the risk of death decreases, which is surprising as high blood pressure usually indicates severe health risks.

The Albumin levels graph includes negative albumin levels, which are impossible, and shows an initial increase in predicted death risk with increasing albumin levels up to about 2.1, contrary to typical understanding that lower albumin levels signify poorer health.

The 'Percentage Bands' graph shows a surprising nonlinear pattern, with the risk of death sharply increasing when 'percentage\_bands' is within the range of 42.0 to 62.0. 

The Creatinine Level graph shows a decrease in predicted risk of death as creatinine levels increase beyond approximately 5.0, which is counterintuitive as higher creatinine levels usually indicate worse health outcomes.

The Sodium Levels graph reveals a sharp decrease in risk at very high sodium levels, a peak in risk at levels of 156 to 159, and a lower-than-expected risk at very low sodium levels, all contrary to standard medical expectations.

These patterns may indicate overfitting, unknown confounding factors, data quality issues, or unique dataset characteristics. Further investigation is necessary for accurate clinical decision-making."}
\end{quote}

Directly interfacing with LLMs can lead to unstructured and confusing results. 
For example, the second surprise listed for albumin (``an initial increase in predicted death risk with increasing albumin levels up to about 2.1'', Figure 4A) can be interpreted in multiple ways. Is the LLM referring to the change in mortality risk from 0.0 to 2.1g/dL that is mostly attributable to the effect of missing albumin values at 0.0g/dL, the change in mortality risk from 1.0 to 2.1g/dL that is mostly attributable to the effect of the jump at 1.5 g/dL, or the gradual increase in mortality risk from 1.5 to 2.1 g/dL? 
This ambiguity motivates development of standardized graph-LLM interfaces and methods to verify LLM interpretations prior to presenting them to the user.

\section*{Discussion}
In this study, we have shown that because LLMs are remarkably good at understanding graphs, they are also able to understand and explain glass-box Generalized Additive Models (GAMs) which decompose learning into an additive sum of feature functions represented as graphs. 
Moreover, LLMs are able to use their extensive background knowledge to detect and describe anomalies in the learned feature functions that defy expectations. 
While the experiments and results reported here are preliminary, they encourage future exploration to automate data science with LLMs.

\subsection*{Memorization, Data Pollution, and Hallucinations: What Happens if the Data or Models Trained on the Data Are in the LLM's Train Set?}
One important concern in using LLMs to automate data science is whether the LLM itself was trained on the studied data, papers written about interpreting models trained on that data, or similar datasets that may cause the LLM to hallucinate insights. 
See \cite{nori2023capabilities} for a discussion of this issue on medical datasets. 
In our preliminary experiments reported here, the pneumonia data we use has not been made public, so we are confident that the pneumonia data is not in the LLM's train set. 
However, there have been several papers that discuss interpretable models trained on the pneumonia data \cite{caruana2015intelligible} and the LLM may have been trained on these papers. 
As a result, LLMs hallucinate results when prompts are too open-ended. 

For example, GPT-3.5 hallucinates surprises about pneumonia patients without ever being given a model:

\begin{quote}
    {\bf Question: } ``You are an expert statistician and data scientist. Your task is to interpret global explanations produced by a generalized additive model (GAM). GAMs produce explanations in the form of graphs that contain the effect of a specific input feature. Answer all questions to the best of your ability, taking into consideration your knowledge about the real world. Pay special attention to values or patterns that appear abnormal. It is very important that you alert the user about these potentially surprising aspects.
    
    This model represents outcomes of hospitalized patients with pneumonia. The outcome is in-hospital mortality.
    What surprises do you find? For each surprise, tell me the specific feature, the feature values (bins) that have surprising effects, and rank the surprise on a scale from 0-5 where 0 means unsurprising and 5 means surprising.''

    {\bf GPT-3.5: } ``As an expert statistician and data scientist, I'll interpret the global explanations produced by the generalized additive model (GAM) for the outcomes of hospitalized patients with pneumonia. Let's discuss the surprises:
    
    1. Surprise: Age has a surprising effect on in-hospital mortality.
       Specific Feature: Age
       Feature Values (Bins): The age bins where the effect is surprising are 18-30 and 70-85.
       Ranking: 5/5
       Explanation: The GAM model shows that patients between the ages of 18 to 30 and 70 to 85 have a significantly higher or lower in-hospital mortality rate compared to other age groups. This finding is unexpected as one might assume that older patients would generally have a higher mortality rate. Further investigation is warranted to understand the underlying factors contributing to this surprising effect.
    
    2. Surprise: Length of hospital stay has a surprising effect on in-hospital mortality.
       Specific Feature: Length of Hospital Stay
       Feature Values (Bins): The effect is surprising for patients with a length of stay between 1 to 3 days.
       Ranking: 4/5
       Explanation: The GAM model indicates that patients with a length of stay between 1 to 3 days have a notably higher mortality rate compared to those with slightly shorter or longer stays. This finding is unexpected as shorter hospital stays are often associated with better outcomes, but here, it seems to be the opposite. Further investigation is needed to understand the reasons behind this surprising trend.
    
    3. Surprise: Blood pressure has a surprising effect on in-hospital mortality.
       Specific Feature: Blood Pressure
       Feature Values (Bins): The effect is surprising for patients with extremely high or low blood pressure values.
       Ranking: 5/5
       Explanation: The GAM model reveals that patients with either extremely high or extremely low blood pressure readings have a significantly higher risk of in-hospital mortality compared to patients with more moderate blood pressure values. This unexpected finding contradicts the common belief that extremely high blood pressure is the primary concern.\footnote{In its hallucination, the LLM invented a belief -- it is well known that extremely low blood pressure is linked to in-hospital mortality \cite{ma2011predictors}.} Further analysis is crucial to comprehend the reasons for this surprising relationship.
    
    4. Surprise: Seasonal effect on in-hospital mortality.
       Specific Feature: Season (Time of Year)
       Feature Values (Bins): The effect is surprising during the winter season.
       Ranking: 3/5
       Explanation: The GAM model suggests that in-hospital mortality rates are higher during the winter season compared to other seasons. While it is not entirely surprising to observe some seasonal variations in health outcomes, the magnitude of the effect during the winter is unexpected. This finding warrants further investigation to identify potential factors contributing to increased mortality during this time.
    
    5. Surprise: Smoking status has a surprising effect on in-hospital mortality.
       Specific Feature: Smoking Status
       Feature Values (Bins): The effect is surprising for current smokers.
       Ranking: 4/5
       Explanation: The GAM model indicates that current smokers have a lower mortality rate compared to non-smokers or former smokers. This finding is unexpected as smoking is generally associated with adverse health outcomes, including higher mortality rates. Further analysis is needed to understand the potential underlying factors contributing to this surprising relationship.
    
    Please note that these surprises are based on the statistical analysis of the GAM model's output and might require further investigation and validation with domain experts and additional data to draw definitive conclusions. Interpretation of statistical models should always be done cautiously, considering the complexity of real-world factors that may influence the outcomes.''
\end{quote}

In this example, we never gave the LLM any model to interpret, but the LLM still readily volunteered hallucinated ``surprises'' and gave plausible explanations. 
This behavior is undesirable. % because the LLM hallucinates insights, providing analyses even without a model or dataset. 
GPT-4 correctly provides hypothetical insights in response to this malformed question:
\begin{quote}
    {\bf GPT-4: }``I'm sorry for the confusion, but I'm unable to view or interpret specific graphs or data directly as I don't have access to external databases or the capability to view images or graphs. As an AI text model developed by OpenAI, I work purely on the basis of the text-based inputs and my training data up to September 2021.
     
    However, I can provide general insights on how to interpret generalized additive model (GAM) results.
    
    In GAMs, a separate smooth function is estimated for each predictor. If you look at the plot for each predictor, the y-axis shows the effect on the response after accounting for the average effects of all other predictors. It's important to examine these plots to see if there's an intuitive or medically-explainable link between the predictor and the response.
    
    For example, let's say you're looking at the age of patients as a feature. If you observe that in-hospital mortality increases sharply for very young and very old patients, but decreases for middle-aged patients, this might be surprising (rank 4-5) given that pneumonia is generally more dangerous for older people due to weakened immune systems and potential comorbidities.
    
    However, in the case of variables like the white blood cell count, you might expect a U-shaped relationship with mortality, with both very low and very high values being associated with increased risk. If you instead observe a simple linear relationship, this might be surprising (rank 4-5), as it would contradict our expectations based on medical knowledge.
    
    Please provide more detailed information about the specific features and their effects for a more detailed interpretation. Remember to also look for interactions between variables, as these can sometimes produce surprising results.''
\end{quote}

These responses demonstrate the dangers of data pollution when interfacing with LLMs. 
Despite the serious concern for memorization and pollution from prior literature, we suspect one of the reasons why LLMs are so good at interpreting graphs is because they have read a vast literature (larger than any one human being) that includes graphs and descriptions of graphs, which has given them significant expertise in interpreting graphs. 
This skill makes LLMs well-matched to additive models such as GAMs that learn models consisting of graphs.

\subsection*{Can LLMs Explain Black-Box Models?}
While we suspect it is unlikely that the current generation of LLMs could be prompted in such a way as to directly understand and explain complex black-box models such as deep neural nets, or large ensembles of boosted trees or random forests, intermediate steps of model explanation could be used to enable LLMs to explain black-box models. 
%For example, by ingesting the output of model explanation methods such as LIME \cite{ribeiro2016should} or SHAP . 
For example, SHAP \cite{lundberg2017unified} generates explanations that are additive and composed of graphs, and can be viewed as a very specific form of GAM trained via distillation on the original black-box model \cite{bordt2023shapley}. 
While SHAP explanations are useful in data analysis, so LLM explanations of SHAP explanations will likely also be useful, it is important to keep in mind that the explanations generated by black-box interpretation methods are approximate explanations of more complex black-box models, and thus any LLM explanations will also be approximate. As a result, strange behaviors could emerge, including instability or adversarial attacks \cite{bordt2022post}. 
In contrast, glass-box models such as EBMs provide exactly additive learned functions, so there is no approximation needed to interpret the model.

\subsection*{Dealing with Finite Context Window Length}
The textual description length of the entire GAM on the pneumonia dataset, after simplifying the graphs by removing small artifacts and rounding numbers to significant digits, is 43,592 GPT-4 tokens. 
Even this relatively simple model contains too many tokens to be directly encoded in GPT-4's 32k context window. 
Black-box models might be much more complex. 
On the other hand, univariate feature graphs are much more compact: the maximum description length of a single graph in the pneumonia GAM is 2,345 GPT-4 tokens. 
This easily fits within GPT-4's context window.

The challenge of breaking models into compact context windows is analogous to the challenge of human interpretability of machine learning models: humans are not good at understanding multiple simultaneous effects. 
However, when effects can be represented as separable and hierarchical components, even complicated effects can be reliably interpreted by humans. 
Because additive models decompose learned functions into an additive sum of terms, both humans and LLMs can effectively reason about GAMs one term at a time. 

\subsection*{The Importance of Separable Components: Graph Scale and Vertical Offset affect LLM Interpretation}
LLMs are sensitive to small variations in how the graphs are presented, including variations that humans would ignore. 
Sometimes this sensitivity is an advantage (e.g. the LLMs can identify jumps in the graph that might be ignored by humans), but other times this sensitivity causes fragility in explanations.
For example, in EBMs, by default each graph is centered vertically so that the mean predicted score from that graph is zero on the train set.  
This vertical offset is arbitrary: because GAM terms are additive, we can shift one graph up (or down) as long as we shift other graphs down (or up) the same amount so that the additive offsets compensate for each other.  

Typically, EBMs set the mean of each component to zero and collect the offsets as an intercept term that reflects the baseline prediction of the model for the entire population. 
However, different choices of centering can be made (e.g. some experts in some domains prefer that the minimum of each graph starts at zero so that the contribution of each graph is always positive or zero). 
Quick experiments suggest that LLMs yield somewhat different results when asked to explain graphs that are always positive vs. graphs that are both positive and negative --- when explaining graphs that have both positive and negative regions, the LLM often focuses on describing regions as adding or subtracting from risk (i.e., positive or negative values), at the expense of focusing on the relative shape of the graph which is unchanged by shifting the graph up or down.\footnote{Interestingly, we have observed this same effect in human experts when they are new to interpreting EBM plots that are mean-centered!} 
%In this paper we vertically offset each graph so that it starts at zero to help the LLM focus on the shape of each graph instead of regions that add or subtract from risk.

\subsection*{GPT-4 vs. GPT-3.5} For this project, we experimented both with GPT-4 and GPT-3.5. We found that GPT-4 is able to work with graphs at a remarkable level of accuracy, whereas GPT-3.5 often struggles to perform the basic tasks of our setup. Consider, for example, this response of GPT-3.5 when we ask the model about the average risk of a 82-year old individual, which can be compared to the reported results. 

\begin{quote}{\bf GPT-3.5:} {\it "Based on the graph, the average risk of a patient aged 82 years old dying from pneumonia would be around 0.264 (or -0.264 in log-odds). This value is obtained by taking the mean value for the age interval that includes 82 years old, which is between 80.5 and 81.5 years old, and is equal to 0.238."}
\end{quote}

Of course, there is a potential that results with GPT-3.5 and other LLMs might be improved given sufficient prompt engineering or fine-tuning. We leave this exploration for future work. 

% \subsection*{Towards Automating Data Science}
% Until recently, most data science consisted of a data scientist training models on data, and then working with human experts to determine if the models were safe and efficacious for deployment or if there were issues that need attention/correction prior to deployment. 
% The combination of LLMs and interpretable glass-box models such as GAMs may allow the LLM to replace the data scientist so that domain experts are able to interact directly with their data and models trained on their data with the LLM serving as an informed intermediary.

\section*{Materials and methods}
\subsection*{Datasets}

\paragraph{Pneumonia}
The 1989 MedisGroups Comparative Hospital Database (MCHD) pneumonia dataset \cite{cooper1997evaluation} contains information on inpatients from 78 hospitals in 23 states in the US between July 1987 and December 1988. 
The MCHD contains over 250 pieces of clinical information that include patient demographic characteristics, history and physical examination findings, and laboratory and radiological results, from which 46 variables were selected \cite{cooper1997evaluation} with known or postulated association with mortality in patients with community-acquired pneumonia. 
We used patient data that were collected during the first 48 hr of hospitalization. 

\subsection*{Methods}
\paragraph{Model Estimation}
We use Explainable Boosting Machines (EBMs), a form of GAM \cite{hastie1990generalized} trained with boosted decision trees \cite{breiman1996bagging,breiman2017classification} using the open-source package \texttt{InterpretML} \cite{nori2019interpretml}. % which was originally design for modeling healthcare data. 
GAMs are ideal to build glass-box models of patient risk because: (1) GAMs can be precisely decomposed into risk curves of single variables for interpretation, (2) the flexible component functions allow risk curves of any shape without any implicit preferences, (3) many treatment protocols and clinical decisions (which sum multiple sources of evidence) are inherently additive models (e.g. SAPS II \cite{le1993new}), APACHE II \cite{larvin1989apache}), (4) GAMs provide the ability to edit the model \cite{wang2021gam} and reason about changes to univariable treatment protocol thresholds. 
We use boosted trees to train GAMs because tree-based models are scale invariant allowing features to be represented in their original natural units (including Boolean, nominal, ordinal, or continuous-valued attributes) without biases of pre-processing. 
Tree-based models can estimate discontinuities that smoother models such as spline-based GAMs and neural networks miss. 
The EBM GAMs in \texttt{InterpretML} are particularly useful for healthcare because they use a round-robin fitting procedure that helps ensure hidden effects are observable in the estimated model. 
The only change we make to the default hyperparameters is to increase the number of rounds of bootstrap sampling to 100, following the convention suggested by the algorithm designers \cite{caruana2015intelligible}. 
These GAMs also provide state-of-the-art accuracy on tabular data (benchmarked in Table S1 of \cite{lengerich2022death}).

\subsection*{Code availability}
A \texttt{Python} tool for LLM-GAM interface, automated model analysis, and surprise finding is available at: \newline \packagelink. 

%\section*{Acknowledgements}

\clearpage
%\bibliographystyle{unsrt}
%\bibliography{talk_to_ebm}

\clearpage
\appendix
\setcounter{table}{0}
\setcounter{figure}{0}
\setcounter{page}{1}
\renewcommand{\thepage}{S\arabic{page}} 
\renewcommand{\thesection}{S\arabic{section}}  
\renewcommand{\thetable}{S\arabic{table}}  
\renewcommand{\thefigure}{S\arabic{figure}}

\noindent
\textbf{Supplemental Information for:\\ \emph{\mbox{\titlephrase}}}

\begin{figure}[ht]
    \centering
    \includegraphics[width=0.8\textwidth]{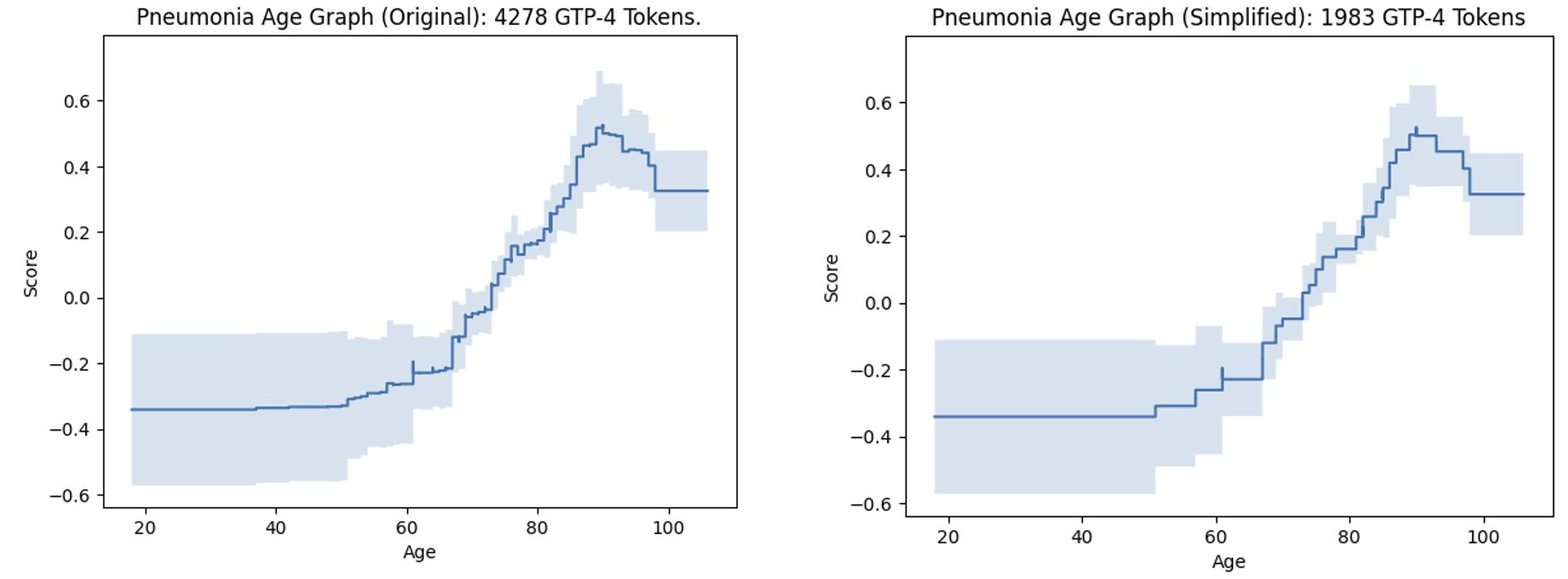}
    \caption{When a single graph is already very complex, we simplify the graph. Often, this allows us to significantly reduce the number of tokens without any significant effect on the shape of the graph. Here, we were looking for a representation of a graph with at most 2000 GPT-4 tokens. }
    \label{fig:climate_graph}
\end{figure}

\begin{figure}[htp]
    \centering
    \includegraphics[width=\textwidth]{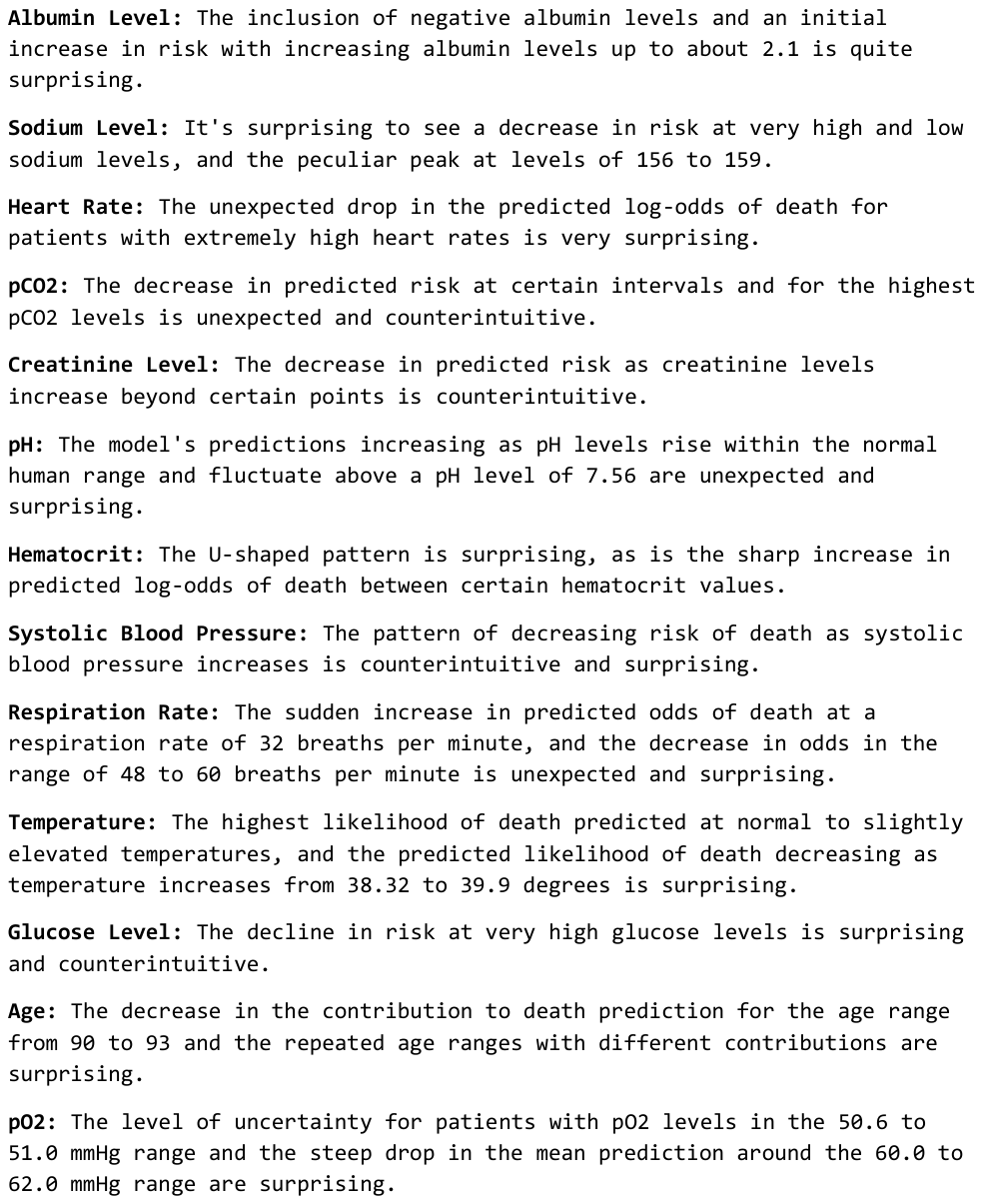}
    \vspace{-10em}
    \caption{The LLM can summarize the surprising effects in many different ways. Here the GPT-4 provides very brief summaries of the most important surprises (importance is judged by the model itself).}
    \label{fig:pneumonia_surprises_short}
\end{figure}

\end{document}